\documentclass[letterpaper, 10 pt, journal]{IEEEtran}
\usepackage{amsmath, amssymb, amsfonts}
\usepackage{algorithmic}
\usepackage{algorithm}
\usepackage{array}
\usepackage[caption=false,font=normalsize,labelfont=sf,textfont=sf]{subfig}
\usepackage{textcomp}
\usepackage{stfloats}
\usepackage{url}
\usepackage{verbatim}
\usepackage{graphicx}
\usepackage{cite}
\usepackage{adjustbox}
\usepackage{diagbox}
\usepackage{caption}
\usepackage{authblk}

\renewcommand{\footnoterule}{
  \kern -3pt
  \hrule width 100pt height 1pt
  \kern 2pt
}

\begin{document}

\title{Enabling Mammography with Co-Robotic Ultrasound}
\author{Yuxin Chen$^\dagger$$^{1}$, Yifan Yin$^\dagger$$^{1}$, Julian Brown$^{1}$, Kevin Wang$^{1}$, Yi Wang$^{1}$, Ziyi Wang$^{1}$, \\ Russell H. Taylor$^{1}$ \IEEEmembership{Life Fellow, IEEE}, Yixuan Wu$^{1,2}$ \IEEEmembership{Member, IEEE}, Emad M. Boctor$^{1,3}$ \IEEEmembership{Senior Member, IEEE} \thanks{$^\dagger$Equal contribution, alphabetically ordered} \thanks{$^{1}$All authors are with \textit{the Laboratory of Computational Sensing and Robotics} (LCSR) at the Johns Hopkins University, Baltimore MD, USA.} \thanks{$^{2}$ywu115@jhu.edu; $^{3}$eboctor@jhu.edu}}

        % <-this % stops a space




\maketitle

\begin{abstract}

Ultrasound (US) imaging is a vital adjunct to mammography in breast cancer screening and diagnosis, but its reliance on hand-held transducers often lacks repeatability and heavily depends on sonographers’ skills. Integrating US systems from different vendors further complicates clinical standards and workflows. This research introduces a co-robotic US platform for repeatable, accurate, and vendor-independent breast US image acquisition. The platform can autonomously perform 3D volume scans or swiftly acquire real-time 2D images of suspicious lesions. Utilizing a Universal Robot UR5 with an RGB camera, a force sensor, and an L7-4 linear array transducer, the system achieves autonomous navigation, motion control, and image acquisition. The calibrations, including camera-mammogram, robot-camera, and robot-US, were rigorously conducted and validated. Governed by a PID force control, the robot-held transducer maintains a constant contact force with the compression plate during the scan for safety and patient comfort. The framework was validated on a lesion-mimicking phantom. Our results indicate that the developed co-robotic US platform promises to enhance the precision and repeatability of breast cancer screening and diagnosis. Additionally, the platform offers straightforward integration into most mammographic devices to ensure vendor-independence.

\end{abstract}

\begin{IEEEkeywords}
breast cancer, co-robotic ultrasound, mammography, autonomous imaging, motion control
\end{IEEEkeywords}

\section{Introduction}
\label{sec:introduction}

Breast cancer has become the most predominant malignancy globally, with 2.3 million incidents and 685,000 deaths in 2020 \cite{arnold2022current}. % Its death rate for female in the United States was 19.6 per 100,000 women\cite{SEER}. 
To mitigate breast cancer fatalities, early detection plays an important role. Among many detection methods, mammography has been the gold standard screening for decades. Across the United States, there were over 27.3 million mammography screenings done in 2019\cite{santo2023national}. However, mammography has its intrinsic limitations, especially its sensitivity for dense breasts. A study in Denmark found that the sensitivities for women with Breast Imaging Reporting and Data System (BI-RADS) density scores of 2, 3, and 4 were only 67\%, 61\%, and 41\%, respectively \cite{von2019sensitivity}.

% \textcolor{green}{US in Breast Cancer Screening}

It is known that US exhibits high sensitivity whereas mammography presents high specificity, in consequence the combination of the two can improve the sensitivity of breast cancer detection overall\cite{kelly2010breast}. Given that US imaging is cost-efficient, widely available, and real-time, the integration of US with mammography yields relatively low financial and temporal outlay. Nevertheless, training and regulation of additional medical personnel causes an appreciable increment in labor costs. Also, human operators will lead to variability in image quality and difficulty in locating the exact imaging plane.

In recent years, researchers have sought co-robotic US to overcome these limitations \cite{vitrani2015robot, li2021overview, boctor2019robot}. One of the innovations is the handover US robotic system. In this setup, operators can guide a robot-held transducer to enable manual control with robotic precision \cite{costanzo2021handover}. This hybrid approach facilitates real-time adjustments during scanning, ensuring accurate localization of lesions while maintaining human oversight. This is particularly advantageous in complex or sensitive imaging scenarios. However, the handover system still requires a level of skill and training, which may limit its accessibility in some clinical settings \cite{costanzo2021handover}.

On a different note, The emergence of automated breast ultrasound (ABUS) in 2012 marked a significant stride in breast imaging, especially as a supplemental screening tool for women with heterogeneously and extremely dense breasts \cite{boca2021pros}. Utilizing an automated transducer, ABUS captures 3D volumetric images, providing a thorough view of the breast tissue. While ABUS excels in comprehensive scanning, it is not designed for specific lesion localization, which may necessitate longer imaging time and additional imaging or complementary use of other imaging modalities for precise lesion characterization and localization \cite{boca2021pros, allajbeu2021automated, nicosia2020automatic}. In addition, it is a dedicated breast US system which is not suitable for other anatomies.

Building upon the foundation of handover US robotic system and ABUS, the field is advancing with autonomous co-robotic US platforms. These systems leverage precise robotic manipulators to autonomously navigate and adjust the US probe, thereby significantly reducing operator dependency \cite{welleweerd2020design}. The automated nature of these systems not only improves the repeatability but also enhances the accuracy of lesion localization. They offer a higher degree of customization and can be programmed to consistently follow predefined scanning protocols. Moreover, the integration of advanced image processing and machine learning algorithms further bolsters the diagnostic accuracy and real-time lesion localization capabilities of these systems \cite{welleweerd2020design, xie2020stabilized}, and greatly enables quantitative US imaging where accurate position of the instrument must be known\cite{deeba2021multiparametric, cloutier2021quantitative, gilboy2020dual, aalamifar2015co}.

Even though co-robotic US introduces additional cost on the purchase and maintenance of a robot arm, it is an economic solution overall due to reduced expenses on training and human resources \cite{dietrich2019editorial, hidalgo2023current, advincula2007robot}. The cost of an industrial robot arm is typically in the range of \$20k to \$50k nowadays. Furthermore, unlike alternative secondary breast imaging methods such as MRI, photoacoustics imaging, elastography, or US tomography that require high-cost or stand-alone systems, the co-robotic US platform offers a vendor-independent solution that can be integrated to most mammographic devices \cite{zhu2020review, bamber1979ultrasonic}, potentially reducing the cost and minimizing the complication to the current clinical workflow.

The paramount importance for a co-robotic US system is the safety, as patients are in direct contact with the moving robot tool and may be in a vulnerable state. Force-based strategies are commonly employed in ensuring safety. These approaches involve the use of external or integrated force/torque sensors in the robot and impedance/admittance-controlled motion modes. Such techniques contribute to safer human-robot interactions by allowing for sensing and mimicking the behavior of a multidimensional spring-damper system \cite{von2021medical, fang2017force, finocchi2017co}. In addition to force-based approaches, other types of safety strategy can be filtering haptic commands and reducing velocity for improved safety, as demonstrated with the UR5 robot \cite{geng2020study}. Some systems, like the ProSix C4 robot, rely on visual inspection by the operator via a camera for safety and surveillance during US image acquisition \cite{huang2019remote}.

In this work, we develop a co-robotic breast US imaging platform. We demonstrate the feasibility of integrating co-robotic breast US to most existing mammographic devices, given that the compression of breast to around 4 cm in mammography is a suitable imaging depth for US and allows a US transducer to scan from the compression plate side. The platform combines a UR5 robot arm, an RGB camera, a force sensor, and an L7-4 linear array for safe and autonomous navigation, motion control, and image acquisition. By addressing the cost and the complication of current secondary breast imaging modalities due to the inclusion of stand-alone or customized imaging systems, our work aims to achieve a vendor-independent execution that can be easily adopted to the current clinical workflow, so that more detailed and accurate information of breast tissue can be provided without increasing the burden on patients or healthcare providers.

\section{Methods}
\iffalse
\begin{figure}[h]
\centerline{\includegraphics[width=\columnwidth]{Figures/loop-x.jpg}}
\caption{The Loop-X X-ray machine}
\label{fig:loopx}
\end{figure}
\fi

\subsection{System Overview}

\begin{figure}[h]
\centerline{\includegraphics[scale=0.27]{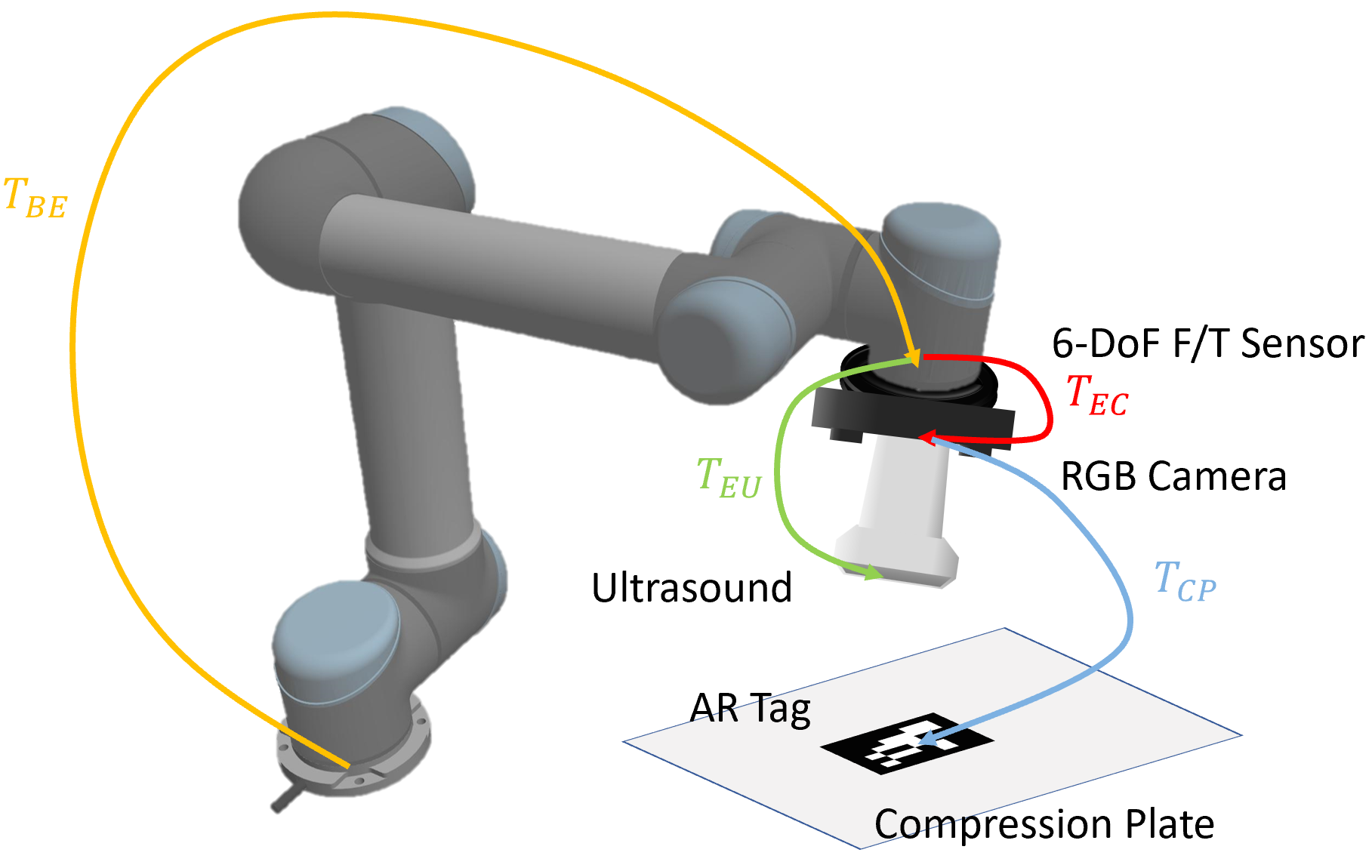}}
\caption{A diagram of the co-robotic mammography system and the transformations between components. T denotes the homogeneous transformation. The components indicated in the subscripts are B – robot base, E – robot end-effector, C –camera, U – US image, and P – mammographic device (the compression plate).}
\label{fig:system}
\end{figure}

To autonomously acquire US images that are registered with mammograms, we utilized a robot arm to hold the US transducer. We equipped the robot arm with a RGB camera (Point Grey, Chameleon, Teledyne FLIR LLC, U.S.) as "eyes" to enable the robot arm with vision. We also attached Aruco AR tags which are both visible in the RGB camera and X-ray images to let the robot arm locate the mammographic device. As shown in Fig. \ref{fig:system}, several calibrations are performed beforehand, where the transformations of interest are 
$T_{BE}$ from the UR5 base to the UR5 end effector, $T_{EC}$ from the UR5 end effector to the camera, $T_{EU}$ from the UR5 end effector to US image, $T_{CU}$ from the camera to US image, and $T_{CM}$ from the camera to the mammographic device (the compression plate), respectively. 

\begin{figure}[h]
\centerline{\includegraphics[width=\columnwidth]{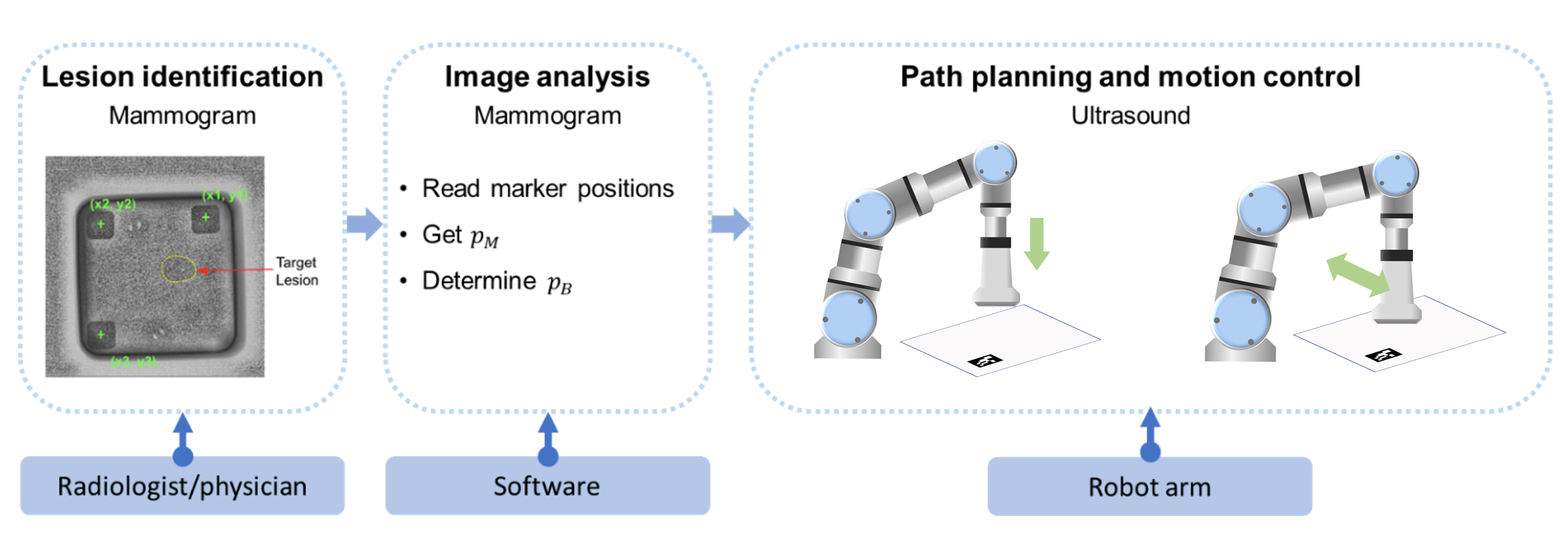}}
\caption{A schematic of the co-robotic mammography process. X-ray images are acquired using an integrated mammography device, and target lesions are identified by radiologists or physicians. Image analysis with dual-modal AR markers on the compression plate calculates the US probe's pose at target locations. A path planning algorithm guides the robot arm to target poses, and a linear scan acquires a small volume of US images while maintaining a constant contact force between the transducer and the compression plate.}
\label{fig:pipeline}
\end{figure}

The co-robotic US interface can generally work in two different schemes. One is automatic 3D US imaging of the whole breast volume. Given that 3D breast US has been well documented as in ABUS \cite{boca2021pros}, we will not discuss it in this work in detail. The other scheme is to acquire slices of 2D US images of specific suspicious lesions only to reduce the overall screening or diagnosis time. The proposed operating pipeline is shown in Fig. \ref{fig:pipeline}. In this study, Loop-X (BrainLAB, IL, USA) was used to take 2D top view X-ray images of the phantom, which is identical to the cranial caudal view in mammography. Sonic Touch (Ultrasonix, Canada) and an L7-4 linear array transducer were employed to acquire US images.

In the first step, we capture an X-ray image of the dual modality phantom using Loop-X. Then an operator marks the center of any visible target lesion, while the co-robotic interface detects the centers of AR markers. Specifically, the RGB camera mounted on the robot arm identifies the phantom's location through patterns of four AR markers. Once we know the marker centers from both the X-ray and the camera, we calculate the transformation to determine the target's 3D position. Afterwards, the robot moves to this position. An automatic scan will follow to obtain a 3D image around the lesion for subsequent analysis.

\begin{figure}[h]
\centerline{\includegraphics[width=\columnwidth]{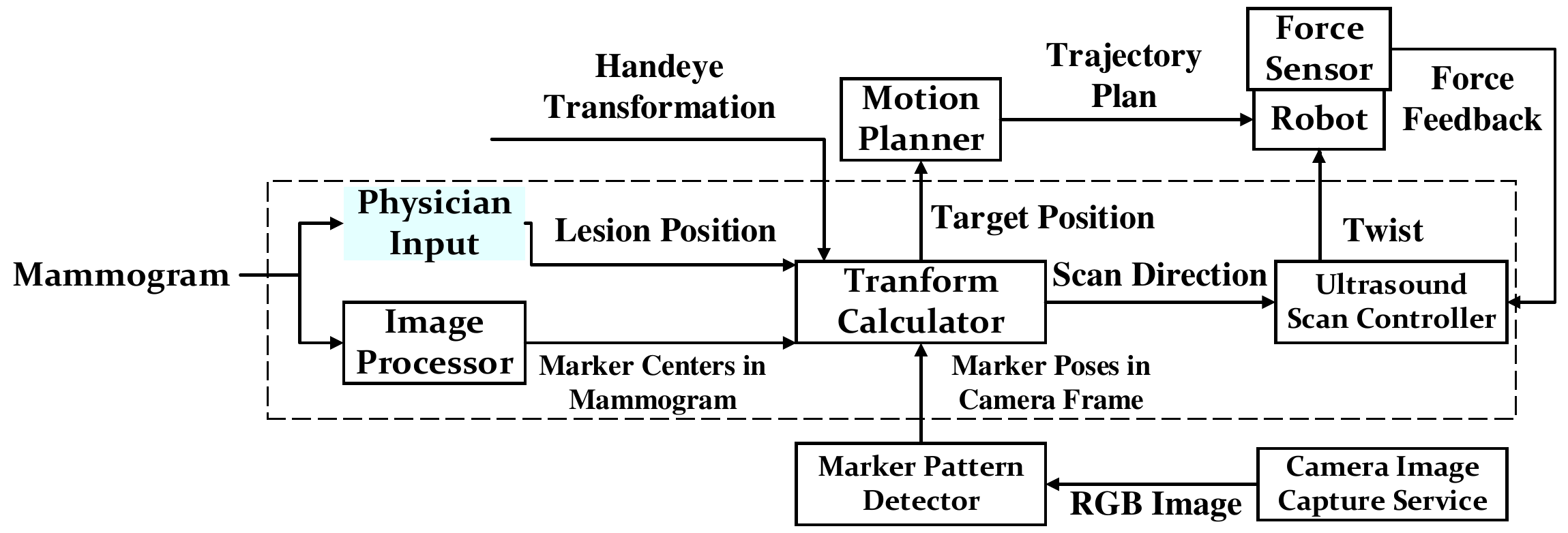}}
\caption{A block diagram of the co-robotic interface. The full pipeline is integrated into a driver ROS node and is shown by a box in dashed lines. The node takes physician's manual annotation of suspicious lesions in the mammogram as input. Then, the Image Processor detects the centers of AR markers, meanwhile Marker Pattern Detector reports the poses of these markers in the camera coordinate. Based on the marker poses, Transform Calculator calculates the target transducer position in the world frame. The Motion Planner will be called right after to plan a trajectory in the joint space. Thereafter, the transform calculator calculates the scan direction of the US transducer and passes it to US Scan Controller. The controller receives force data to control the robot in a Cartesian space control mode by sending twist messages throughout the scanning process.}
\label{fig:block_diagram}
\end{figure}

The co-robotic framework was developed in Robot Operating System (ROS) . Fig.~\ref{fig:block_diagram} illustrates the software architecture based on the data flow among software modules and hardware. 

\subsection{Calibration}

\subsubsection{Camera Calibration}
\begin{figure}[h]
\centerline{\includegraphics[scale=0.6]{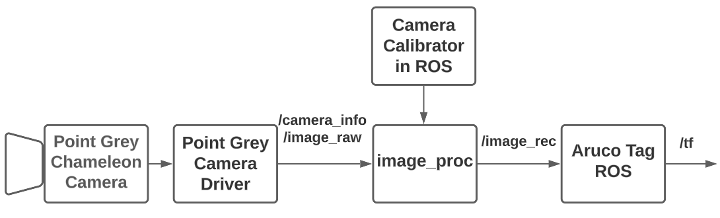}}
\caption{Flowchart of the camera setup and calibration. \emph{image\_proc} is a ROS package that generates rectified camera images. camera\_info is the rostopic that includes the information of the camera matrix, image\_raw is the rostopic that includes the raw image from the camera, and image\_rec is the rostopic that shows the camera image after rectified and calibrated based on the information from camera\_info using image\_proc.}
\label{fig:camera_flowchart}
\end{figure}

We mounted an RGB camera at the end-effector of the robot arm as indicated in Fig. \ref{fig:system}. We employed the Camera Calibrator driving module in ROS for automatic calibration and a 40 cm by 30 cm checkerboard with 8 by 6 squares to provide patterns. The rectification matrix and camera matrix were exported through the \emph{image\_proc} ROS package to generate the rectified camera images. A detailed camera calibration flowchart is shown in Fig.~\ref{fig:camera_flowchart}.

\subsubsection{Hand-eye calibration}
In short, hand-eye calibration solves an $AX=XB$ problem that finds the transformation $T_{EC}$ from the end-effector of UR5 to the camera \cite{tsai1989new}. Here, let $T_{BE_{i}}$ be the transformation from UR5’s base to the end-effector for pose $i$ ($i = 1, 2, \dots, N$), where $N$ is the total number of poses, $T_{CT_{i}}$ be the transformation from the camera to the marker (AR Tag) for pose $i$. As shown in Fig.~\ref{fig:system}, we have: 
\begin{equation}
T_{BE_{1}}T_{EC}T_{CT_{1}} = T_{BE_{2}}T_{EC}T_{CT_{2}}
\end{equation}
Rearrange the equation in the form of $AX=XB$:

\begin{equation}
T_{BE_{1}}^{-1}T_{BE_{2}}T_{EC} = T_{EC}T_{CT_{1}}T_{CT_{2}}^{-1}
\end{equation}

where $X = T_{EC}$, $A = T_{BE_{1}}^{-1}T_{BE_{2}}$, and $B = T_{CT_{1}}T_{CT_{2}}^{-1}$. $AX = XB$ can be solved by using the Kronecker product method \cite{park1994robot}.

\subsubsection{Ultrasound Calibration}
US calibration aims to determine the transformation matrix from the end-effector of UR5 to the US image, and it can be achieved by solving a $B_iXp_i = B_jXp_j$ problem \cite{mercier2005review}. Let $B = T_{BE}$, $X = T_{EU}$, and $p$ be the location of the point fiducial in the US image frame. By fixing the point fiducial with respect to the UR5 base frame and changing the robot pose, for each pair of poses $i$ and $j$, we have: 
\begin{equation}
B_iXp_i = B_jXp_j
\end{equation}
To solve for the $BXp$ problem, a gradient descent algorithm is applied \cite{ruder2016overview}. 

\begin{figure}[h]
\centerline{\includegraphics[scale=0.3]{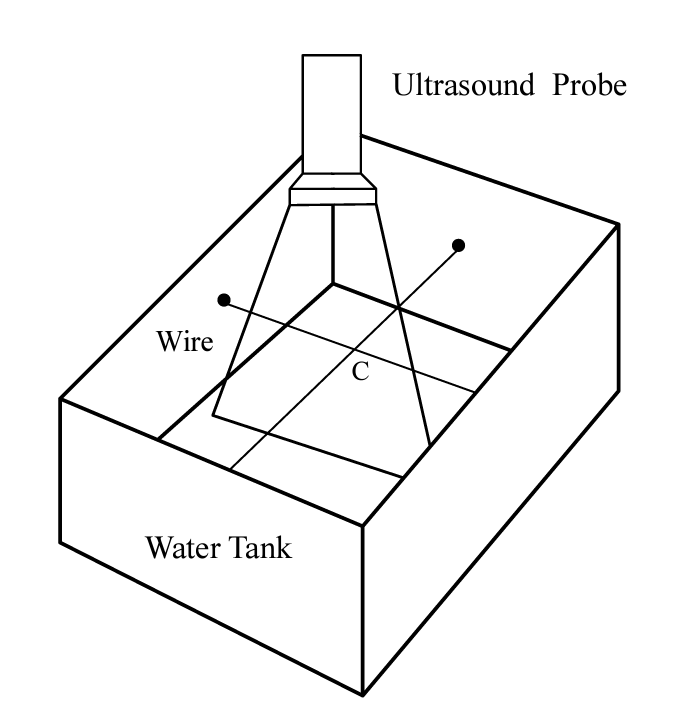}}
\caption{A schematic of the fishing wire cross-wire phantom.}
\label{fig:cross_wire_phantom}
\end{figure}

We built a cross-wire phantom with a water tank for collecting calibration data as depicted in Fig.~\ref{fig:cross_wire_phantom}. The intersection of the cross-wire appears as a point target only if it is in the US image plane. Radio Frequency (RF) US data were collected at different robot configurations. Previous research has demonstrated that this method yields highly accurate calibration outcomes \cite{pagoulatos2001fast}.

\subsubsection{Force Sensor Calibration}

Force sensor calibration aims to compensate the  gravitational influence of tools and sensors mounted on a robot's end-effector. By doing so, it ensures that the sensor readings, both force and torque components, register as zero in the absence of external loads. This challenge is approached as a multi-dimensional curve fitting problem utilizing the Bernstein polynomial \cite{lorentz2012bernstein}. Specifically, we sample an ample set of correspondences between robot tool poses (6-DoF) and sensor readings (6-DoF) across the manipulator's operational workspace. A Bernstein polynomial is then fitted via least squares regression, establishing a mapping between the robot's kinematics and the sensor outputs. Once this mapping is constructed, real-time sensor readings are adjusted by subtracting the uncalibrated readings, based on the current pose, from the present loaded readings.

\subsection{Image Analysis}

The center positions of Aruco markers were found in X-ray images and they were used to determine the target position of the robot-held transducer. In this work, because we are only using one RGB camera which does not give us depth information diretly, we are using four AR markers, whose centers form a bounding box. The compression plate was made of TPX in high rigidity so we neglect its deformation. Let the homogeneous representation of the center of a suspicious lesion in the mammogram coordinate be  $p_M^l \in \mathbb{P}^2$ . We are looking for the transformation from $p_M^l$ to the target transducer position $p_B^l \in \mathbb{P}^3$ in the robot base coordinate.

To do so, we first determine the rigid body transformation $T_{BP}$ from the robot base to the compression plate. We utilized the “aruco-ros” package (Aruco-ROS, 2021) to estimate the pose $T_{C{T_i}}$ of Aruco marker $i$ in the camera view, which then can be converted into the robot base coordinate by $T_{B{T_i}} = T_{BC} T_{C{T_i}}$. Assume $p_i$ is the translation component of $T_{B{T_i}}$, 
then the transformation $T_{BP} \in SE(3)$ can be expressed as
\begin{equation}
T_{BP} = \begin{bmatrix}
n_x & n_y & n_z | \frac{1}{4} \sum_{k=1}^4 p_{k}
\end{bmatrix}
\end{equation}
where $\frac{1}{4} \sum_{k=1}^4 p_{k}$ is the center of the bounding box.

A natural choice of the z-axis of its rotation component ($n_z$) is the average of z-bases of all the markers
\begin{equation}
n_z = \frac{\sum_{k=1}^4 n_{z_k}}{\left\lVert \sum_{k=1}^4 n_{z_k} \right\rVert}
\end{equation}

To determine $n_y$, we first find $n_y'$ which is parallel to the edge of the bounding box given by
\begin{equation}
n_y' = \frac{(p_4 - p_1) + (p_3 - p_2)}{\left\lVert (p_4 - p_1) + (p_3 - p_2) \right\rVert}
\end{equation}
Ideally if there is no distortion either on the compression plate or from the camera calibration, $n_y' = n_y$. However, in case of any distortion, we impose another constraint to make sure $n_y$ is perpendicular to $n_z$
\begin{equation}
n_y = \frac{n_y' - (n_y' \cdot n_z) n_z}{\left\lVert n_y' - (n_y' \cdot n_z) n_z \right\rVert}
\end{equation}

The x basis ($n_x$) is calculated by the cross product
\begin{equation}
n_x = n_y \times n_z
\end{equation}

\begin{figure}[h]
\centerline{\includegraphics[width=\columnwidth]{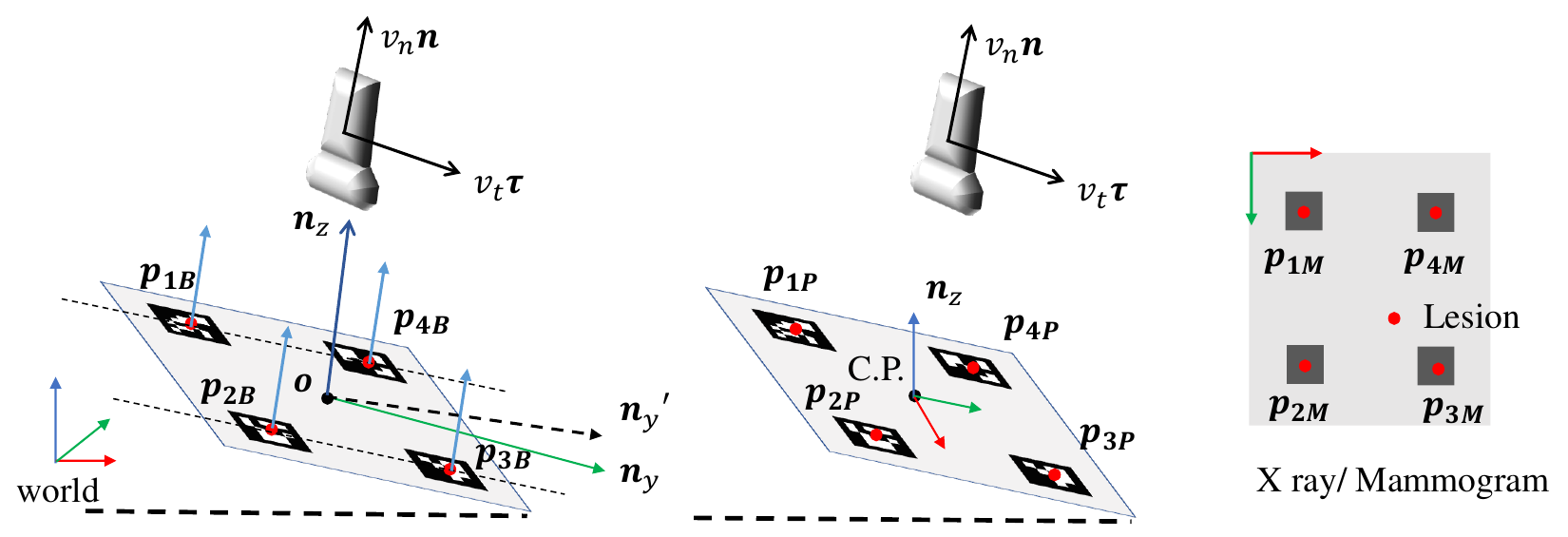}}
\caption{Estimation of transformations using dual-modal markers and force feedback-based motion control. $p_{iB}$ is the marker center position in the robot base frame, $p_{iP}$ is the same point in the frame of the compression plate, $p_{iM}$ is the corresponding image coordinate in the x-ray. 
Shown as two vectors on the probe, the force control loop adjusts a small normal component of scanning velocity based on force/torque sensor feedback, while maintaining a constant tangential velocity.}
\label{fig:image_analysis}
\end{figure}

We then determine a 2D homographic transformation $H_{PM} \in \mathbb{R}^{3 \times 3}$ that involves translation, rotation, affine transformation, and projection from the plane of the compression plate to the mammogram.

The four marker centers in the compression plate frame are calculated as
\begin{equation}
p_{iP} = (T_{BP})^{-1} p_{iB}
\end{equation}
By setting z coordinate of $p_{iP}$ to zero, s.t. $p_{iP}' = \begin{bmatrix} x_i & y_i & 1 \end{bmatrix}^T$, we obtain four correspondences $p_{iP}' \in \mathbb{P}^2 \leftrightarrow p_{iM} \
\in \mathbb{P}^2$, where $p_{iM}$ is the center of the marker in the x-ray image. The Direct Linear Transform (DLT) \cite{hartley2003multiple} is applied to estimate entries of $H_{PM}$ as the smallest eigenvalue of $A^T A$, where
\begin{equation}
A = \begin{bmatrix}A_1 \\ A_2 \\ A_3 \\ A_4
\end{bmatrix}
\end{equation}
and $A_i$ is given by

\begin{equation}
A_i = \begin{bmatrix}
\mathbf{0}^T & p_{iM}'^T & -y_i p_{iM}'^T \\
-p_{iM}'^T & \mathbf{0}^T & x_i p_{iM}'^T
\end{bmatrix}
\end{equation}

Therefore, for any lesion $p_M^l$ in the mammogram, the target transducer position is given by
\begin{equation}
p_B^l = T_{BP} \begin{bmatrix}
D H_{PM} ~ p_M^l \\ 0 \\ 1
\end{bmatrix}
\end{equation}
$D$ is the dehomogenization matrix
\begin{equation}
D = \begin{bmatrix}
1 & 0 & 0 \\
0 & 1 & 0\\
\end{bmatrix}
\end{equation}

\subsection{Force-feedback Motion control and Path Planning}

We decompose the robot motion for US image acquisition into three steps: navigate the US transducer right above the lesion, move it downward until it contacts the compression plate, and scan.

In the first step, the robot arm positions the US probe above the lesion by a preset offset $k_0$. The target position $p_B^g$ is given by
\begin{equation}p_B^g = p_B^l + k_0n \qquad (13)\end{equation}
where $p_B^l$ is the center position of the lesion in the robot base frame, and $n$ is the normal of the compression plate. Rapidly exploring random tree (RRT) \cite{lavalle1998rapidly} was employed to achieve point-to-point path planning. 

In the second step where the US probe dives down until it contacts the compression plate, we leveraged a Robotiq FT-150 Force Sensor for force feedback and control. The force sensor was mounted between the end-effector and the US probe to provide 6-axis force and torque exerted on the end effector. Sensor output data was published onto a ROS topic through the Robotiq ROS package. To ensure safety as well as acoustic coupling, the downward movement stops until the contact force reaches 2 Newtons.

In the third step, linear scanning is performed to capture a small volume of images for a comprehensive view of the lesion.

During the scanning process, a cartesian velocity control was applied to the robot tool center position using a Proportional-Integral-Derivative (PID) force feedback controller \cite{aastrom2021feedback}. The controller limits the magnitude of the contact force within the threshold value. The scanning velocity vector $v$ is given by
\begin{equation}
v = v_t \tau + v_n n
\end{equation}
where $v_t \tau$ is the velocity component parallel to the plane, and $v_n n$ is the component orthogonal to the plane. We also set $v_t$ to be a constant throughout the scanning. The following control law is applied
\begin{equation}
v_n = K_p \left(e(t) + \frac{1}{T_i} \int_0^t e(\tau) d\tau + T_d \frac{d}{dt} e(t)\right)
\end{equation}
\begin{equation}
e(t) = F^* - F
\end{equation}
where the normal speed $v_n$ is the control input, the difference between the sensed contact force magnitude $F$ and the target $F^*$ is the control error $e$, and $K_p$, $\frac{K_p}{T_i}$, and $K_p T_d$ are control gains. The magnitude of the contact force is given by
\begin{equation}
F = \left\lVert F_x \mathbf{i} + F_y \mathbf{j} + F_z \mathbf{k} \right\rVert
\end{equation}
where $F_x$, $F_y$, and $F_z$ are components along $x$, $y$, and $z$ axes, and $\mathbf{i}$, $\mathbf{j}$, and $\mathbf{k}$ are unit orthogonal basis of the tool frame.

\subsection{Validation}

\begin{figure}[h]
\centerline{\includegraphics[width=\columnwidth]{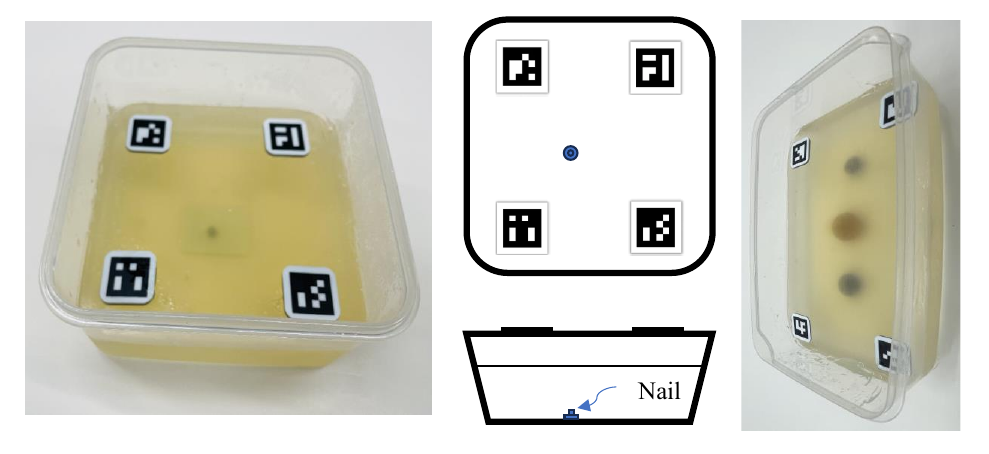}}
\caption{Photos of two dual modality phantoms made of gelatin. A metal nail is placed at the bottom of the first phantom. The top surface of the phantom is covered by a compression plate made of TPX. Four AR markers are places on the top of the compression plate for the localization of the phantom in the experiment. A grape phantom is shown on the right. There is one grape and two blueberries in the phantom and serve as simulated lesions.}
\label{fig:phantom}
\end{figure}

To validate the repeatability of the robotic system, we built a gelatin (Jell-O, IL, USA) phantom to mimic breast tissues. A nail was embedded inside the phantom to provide both strong X-ray contrast and US contrast as shown in Fig.\ref{fig:phantom}. The pinpoint of the nail enables the estimation of the system repeatability. A compression plate was placed on the phantom to mimic realistic mammography setup. The compression plate is made of polymethylpentene or commonly known as TPX (Mitsui Chemicals, Japan), which is a rigid plastic material being optical, X-ray and US transparent. Four Aruco markers are affixed to the compression plate. These markers present distinct patterns in camera images and clear outlines in the X-ray images. We varied the initial positions of the robot arm and measured how accurately it can navigate to the pinpoint of the nail in the phantom. We captured a 3D volume of US images by moving the robot arm along the planned scanning direction in a 0.05 mm increment. The expected target position $p_{exp}$ is where the intensity of the nail shown in the corresponding US image gets the highest value. The mean and standard deviation of the distance between the navigated location and $p_{exp}$ are calculated to evaluate the repeatability.

\begin{figure}[h]
\centerline{\includegraphics[scale=0.5]{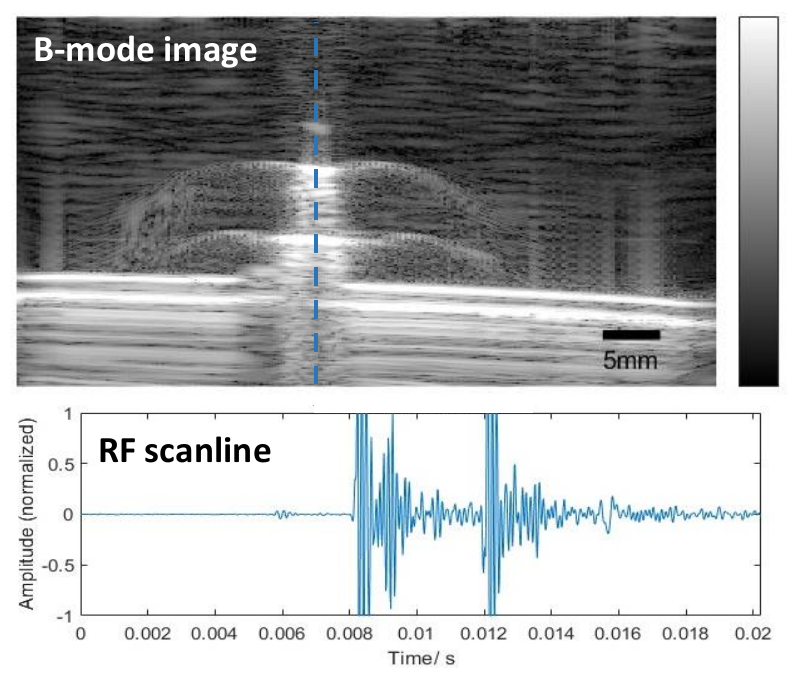}}
\caption{Figure of the B image captured during a trial of the validation experiment. The plot below shows the RF signal along the marked vertical line.}
\label{fig:ultrasound_image}
\end{figure}

To transform the captured RF data into B-mode images, a process was employed that involved taking the magnitude of the analytic signal derived from the RF data using the Hilbert transform and then displaying it in a logarithmic scale. A representative sample of the captured RF data and the resulting B-mode image can be found in Fig.~\ref{fig:ultrasound_image}. This entire process was repeated for ten individual trials to ensure a comprehensive assessment of the system's performance.

To show the feasibility of the whole system and the importance of including US imaging as a secondary modality, we built a fruit phantom with one grape and two blueberries embedded mimicking lesions as shown in Fig.~\ref{fig:phantom}. We specifically chose a grape and two blueberries with high density (they sink in water) to compare their contrast in X-ray and US images. We assessed the Contrast to Noise Ratio (CNR) in both X-ray and US images of the lesion mimicking phantom. The CNR is given by 

\begin{equation}
\text{CNR} = |m_t - m_b| / \sigma_b
\end{equation}
where $m_t$ is the mean signal intensity of the target region, and $m_b$ and $\sigma_b$ are the mean and the standard deviation of the background intensity, respectively.
Additionally, a 3D volume is reconstructed from an ultrasound scan of the phantom, allowing for the comparison of images from the two modalities under the same view.

\section{Results}
\subsection{System Accuracy}

\begin{figure}[h]
\centerline{\includegraphics[scale=0.6]{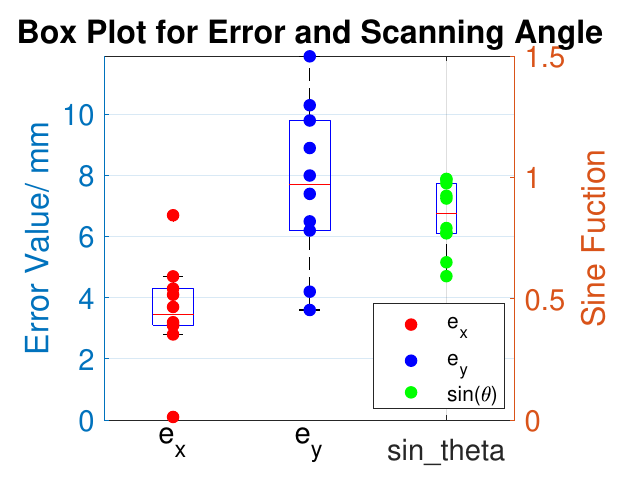}}
\caption{Figure of the box plot for the positional system errors and scanning angles over ten trials.}
\label{fig:box_plot}
\end{figure}

The validation results are illustrated in Fig.~\ref{fig:box_plot}. Our findings indicate that the mean error in the lateral ($e_x$) direction was $3.60\pm1.58$mm. In the axial ($e_y$) direction, the error was $7.70\pm2.51$mm. These results demonstrate the system's capability to consistently position the US probe near a designated target. 

While absolute accuracy and precision are of value, they are not imperative in our context. This is because our approach inherently involves scanning a region around the target, accommodating minor positional variances. 

\subsection{Feasibility of Automatic and Localized Lesion Imaging}

%To evaluate the potential for automated and localized lesion imaging, we embarked on an innovative experiment using a simulated breast phantom. This phantom, crafted from jelly, was embedded with grapes that acted as simulated lesions to mimic real-life complexities. A photo of the phantom is shown in Fig.~\ref{fig:grape_phantom}.

% \begin{figure}[h]
% \centerline{\includegraphics[width=\columnwidth]{Figures/grape_phantom.jpg}}
% \caption{A photo of the grape phantom. There is one grape and two blueberries in the phantom and serve as simulated lesions.}
% \label{fig:grape_phantom}
% \end{figure}

%The automation of the imaging process was designed with clinical accuracy and efficiency in mind. Upon identification of a target lesion in the X-ray image by a physician, our system efficiently located the Robot Tool Center Point (TCP) corresponding to the scanning initiation point. Subsequent motion planning elevated the robot above this position, from which it then descended until contact with the phantom was established. The US image acquisition (or scanning) ensued, guided by a meticulously implemented force feedback control mechanism.

\begin{figure}[h]
\centerline{\includegraphics[scale=0.45]{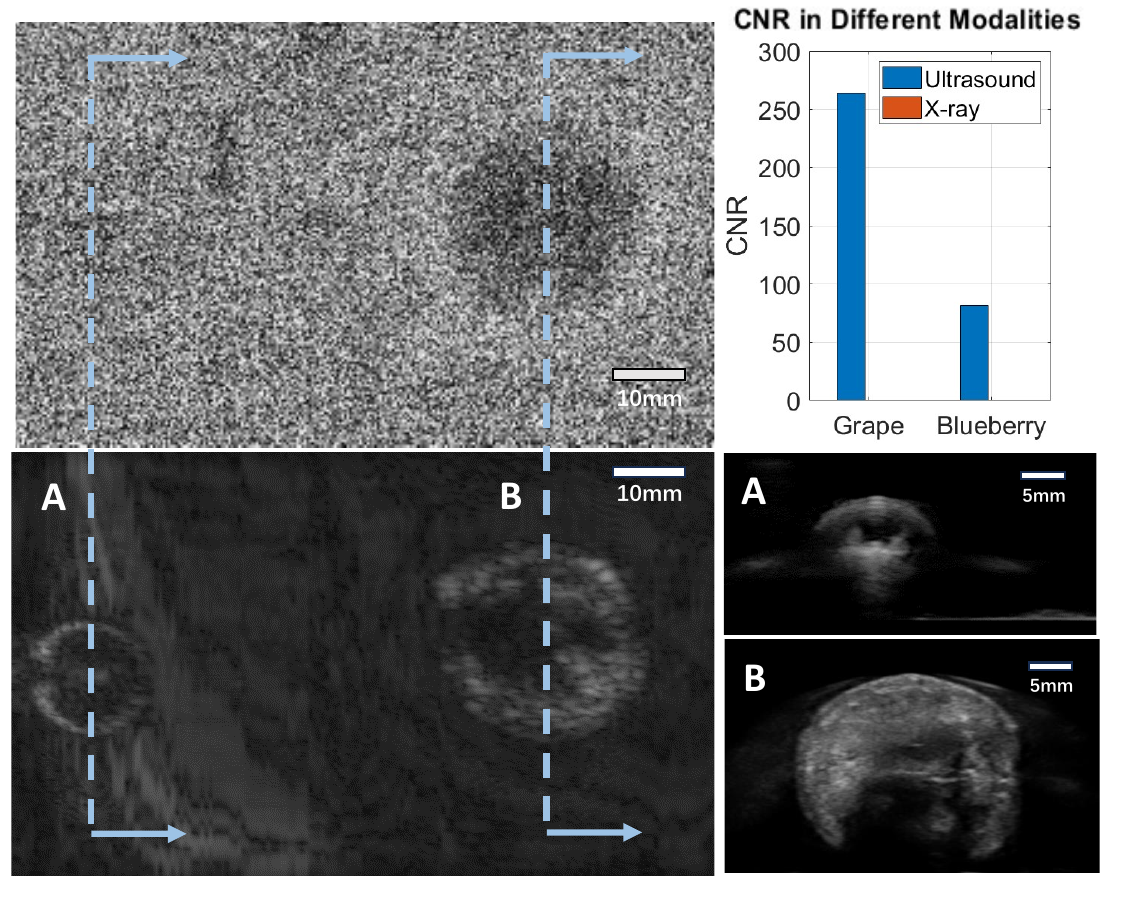}}
\caption{Comparison of the same simulated lesions in both X-ray and ultrasound images, with a barplot depicting the image CNR for each modality (upper right).}
\label{fig:grape}
\end{figure}

We conducted 20 trials on the grape phantom for testing the system efficiency. Each trial started from a different target lesion position and a robot configuration. Our system achieved a $100\%$ success rate in automated image acquisition. The overall procedure took around 1 to 2 minutes on average, which indicates a high efficiency of our co-robotic US interface. A representative X-ray image and the corresponding C-section obtained by slicing the US 3D volume is illustrated in Fig.~\ref{fig:grape}. The actual US images of the two lesions at a specified cross section are displayed in the lower right of the figure.

%\textbf{Qualitative Analysis}

The difference in quality between X-ray and ultrasound (US) images highlights their complementary uses. X-rays offer a basic assessment of lesion location but often lack sufficient contrast, even when imaging a grape phantom. In contrast, US images excel in lesion detection, providing excellent contrast.

% While X-ray images offer a holistic perspective, presenting a broad overview of the entire tissue, US images excel in delivering meticulous vertical details from a central point. This granular representation empowers clinicians with an in-depth understanding of anomalies or lesions, vital for comprehensive diagnosis and treatment planning.

%\textbf{Quantitative Analysis}

We computed the averaged CNR across one X-ray and 171 US images in a volume scanning that contains two simulated lesions (grape and blueberry). The respective CNR values for the X-ray and US images are presented in Fig.~\ref{fig:grape}. Specifically, the CNR for the grape is 264.12 in the ultrasound and 0.46 in the X-ray, while for the blueberry, the CNR values are 81.81 in ultrasound and 0.20 in X-ray.
%The fundamental principles behind each imaging modality play a pivotal role in their CNR values. US, with its reliance on high-frequency sound waves, tends to produce high contrast, especially when imaging fluid-filled structures against solid tissues, such as our jelly phantom embedded with grapes. The reflection and scattering of US waves at tissue interfaces render this high CNR. X-ray imaging, on the other hand, relies on the differential absorption of ionizing radiation in tissues. Given its inherent soft tissue visualization properties, X-rays typically have lower contrast when representing tissues of similar densities.

% In our specific experiment context, focusing on simulated lesions inside a jelly phantom, the US unsurprisingly demonstrated a potentially higher CNR. This can be attributed to the stark contrast differential between the grapes (simulated lesions) and the surrounding jelly matrix. The X-ray, while offering a comprehensive overview of tissue structures, may not pinpoint these lesions with the same contrast depth as the US.

This analysis confirms the complementary nature of both imaging modalities: X-rays presents clear anatomical and morphological structures but the image quality can be limited in dense tissues due to the shadow artifacts. While US delves deeper into tissue properties and gives great contrast even in highly dense tissues.

\subsection{Force Feedback Control}

\begin{figure}[h]
\centerline{\includegraphics[width=\columnwidth]{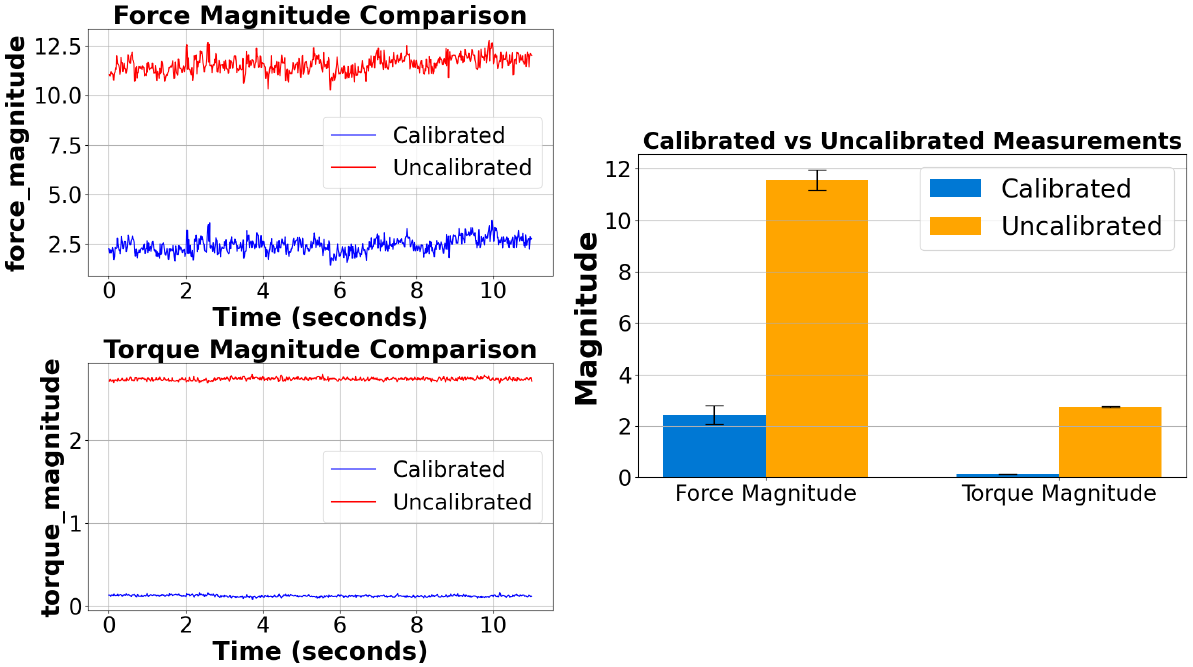}}
\caption{Comparison of force/torque readings: Uncalibrated vs. calibrated over time. Signal-time plot (left) and bar plot representation (right)}
\label{fig:comparison}
\end{figure}

The force sensor calibration results are shown in Fig.~\ref{fig:comparison}. On the left of the figure, temporal profiles of force and torque magnitude from the sensor are displayed, illustrating the difference between pre- and post-calibration readings. The uncalibrated readings, depicted in red, exhibit a high degree of measurement bias and a potential for inaccurate force and torque estimations. In contrast, the calibrated readings, represented by the blue lines, demonstrate output magnitudes much closer to zeros over the 10-second interval. Note that, for the force magnitude, the calibrated reading still has a bias around 2.5N, that is caused by the limited workspace of the configured robot. The robot can’t move freely in every pose when collecting data for the compensation.
On the right, the bar chart comparing calibrated versus uncalibrated measurements is included to provide a statistical synthesis of the sensor's performance in both calibrated and uncalibrated states. Comparing with the uncalibrated, the calibrated force readings (blue bar), exhibit a higher mean value and possess a narrower error bar. This illustrates an improvement in both measurement accuracy and precision through calibration. In the case of torque, the calibrated reading (blue bar) has a much lower mean level around zero, compared to the uncalibrated (orange bar). Also, the calibrated and uncalibrated both have invisible error bars, indicating a small noise level and measurement uncertainty in each case.

\begin{figure}[h]
\centerline{\includegraphics[scale=0.35]{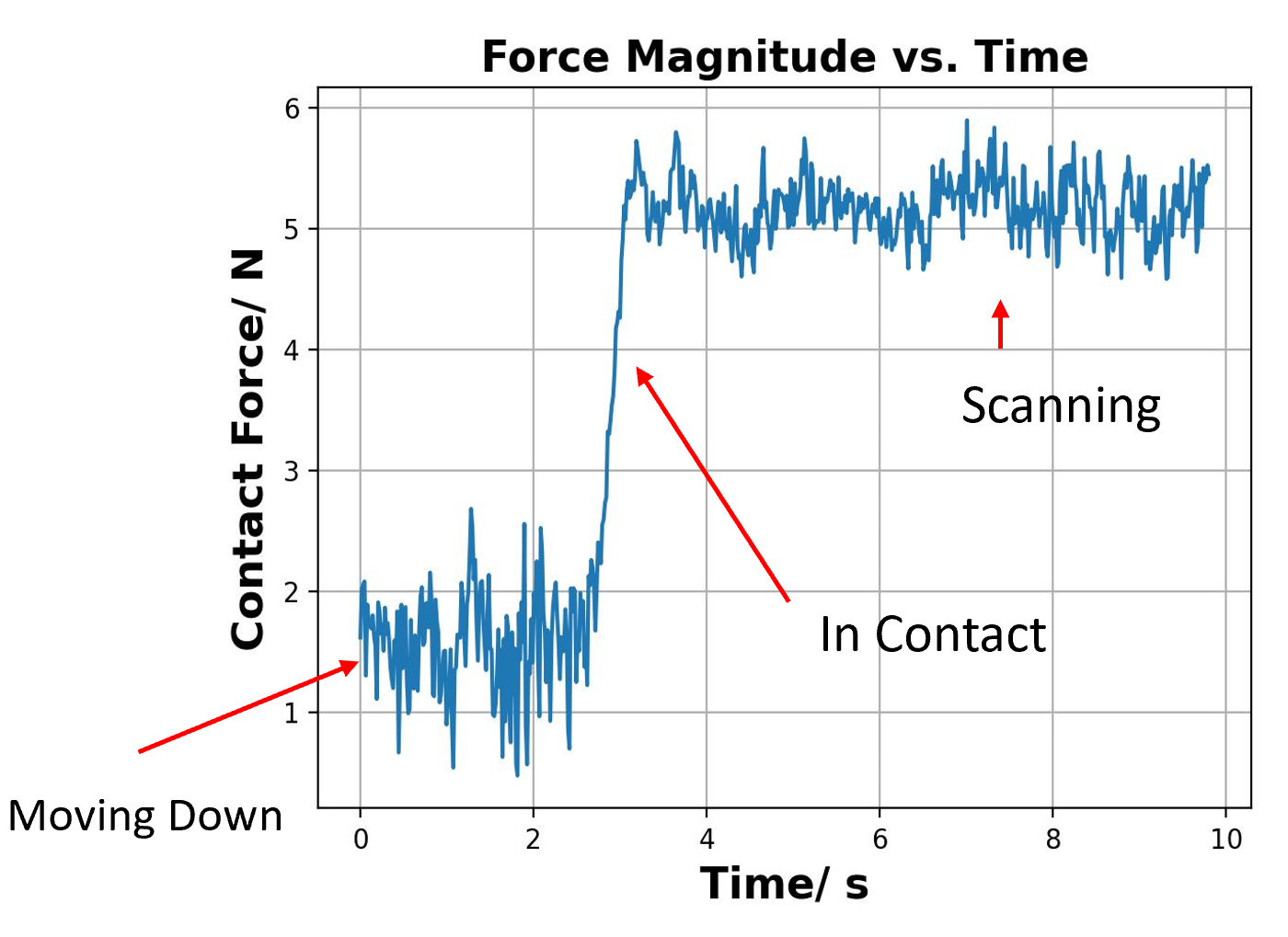}}
\caption{Figure of the force control signal during a full scanning (force magnitude vs. time). It maintains a contact force close to the desired 5N target, with a mean of 5.18N and standard deviation of 0.23N.}
\label{fig:force_feedback}
\end{figure}

In order to evaluate the effectiveness of the force feedback control in our co-robotic US scanning system, an experiment was conducted using the grape phantom. The contact force on the transducer during the entire procedure with respect to time was continuously monitored. The resultant force-time curve is depicted in Fig.~\ref{fig:force_feedback}.

%During the experiment, the probe was descended from above the phantom, 

From Fig.~\ref{fig:force_feedback}, when the transducer was still moving downward in the free air before touching the phantom, the force readings are seen to hover around a minor constant value with small fluctuations due to imperfect force sensor calibration.

As the probe first contacts the phantom, the force profile depicts a relatively linear increase. The linear signal lasts for around 0.4 seconds, followed by the signal fluctuating around a setpoint of 5N, corresponding to the controlled scanning with force feedback control. The overshoot is observed to be 0.72N. The measurements indicate a mean value of 5.18N with a standard deviation of 0.23N. Such consistency shows the feedback control's efficacy in maintaining the desired contact force. The moving speed during scanning is 7.27mm/s. 

The observed stability and precision in the force readings highlight our system's potential to sustain contact force within a safe and comfortable range, ensuring patient comfort throughout the US scanning procedure.

\section{Discussions and Conclusion}
Early detection is crucial in preventing aggressive breast cancer from advancing to later stages and causing death. While many imaging approaches have been developed for breast cancer screening and early diagnosis, they are either not sensitive enough to identify deadly cancer, or overly sensitive but less specific resulting in overdiagnosis and overtreatment. One possible solution is to rely on joint forces, i.e., multimodal imaging, to overcome the trade-off between sensitivity and specificity imposed by individual modalities. However, the inclusion of other modalities faces challenges in costs, amendments of clinical workflows, acquisition of approval from administrations, and popularization. Therefore, there is a growing demand for improved imaging techniques that can provide more detailed and accurate information about breast tissue without increasing the burden on patients or healthcare providers. 

To address the unmet need, in this work, we introduced a vendor-independent, automatic co-robotic US imaging interface as a secondary modality to mammography. We showed the feasibility of integrating the interface to mammographic devices and demonstrated safe, autonomous, and efficient US image acquisition. US images can be acquired either in 3D as in ABUS, or 2D at selected lesion sites for quick checks. Despite minimum adaptation is required to integrate co-robotic US to mammography, it is worth noting that the compression plate must be replaced with echoluscent materials. TPX is a suitable material because it provides rigidity \cite{TPXprop}, optical and X-ray transparency \cite{barmore2019mechanical}, and tissue-like acoustic impedance \cite{madsen2011properties}.

% In contrast to current systems that depend on sonographer expertise and are prone to variability, our automated system minimizes the inherent subjectivity and the potential for human error. It offers a more objective and consistent approach to US image acquisition, addressing some of the limitations in manual methods and potentially paving the way for more consistent and reproducible results.

%\subsection{Limitations}
There are several limitations of our current platform. The repeatability of our co-robotic system is currently at 7 mm. Although a target lesion can always be found by scanning a small volume even if there is an offset between the initial navigated position and the actual target position, a better repeatability allows us to scan a smaller volume or even just one slice of 2D image and thus enables a faster procedure. Two major components in our current setup can be enhanced. One is the target pose estimation through the camera, and the other is US calibration. Instead of using a single RGB camera, target pose estimation could be improved by either adopting a stereo camera or installing two RGB cameras. US calibration accuracy can be enhanced by using more extensive and diverse data points in the BXp problem. It is also possible that errors exist in the kinematic chain of our old UR5 CB2 arm, given that kinematic calibration approach using a calibration board for UR5 CB2 is no longer applicable to newer version arms, and the calibration board is out of stock. We will use a newer version UR5 in the future.

In addition to enhancing the repeatability, we foresee an even faster procedure once the motion control is further optimized. Portability is another concern, given the cumbersome nature of a 6-DoF robot arm. Future designs might consider more compact systems such as UR3 or MECA 500 (Mecademic, Canada). It is worth noting that, while initial costs for a durable robot arm are high, long-term savings could arise from reduced human resource dependence and lower occupational health risks for sonographers.

Future work should focus on testing with a more realistic breast phantom to better replicate clinical conditions, and on exploring multimodality imaging using advanced techniques like US tomography, photoacoustic imaging, and contrast-enhanced imaging. Furthermore, automated breast US using robotics could further minimize manual intervention and enhance efficiency. Validating these advancements will require thorough studies on human cadavers and subsequently on live human subjects to confirm the system's safety, accuracy, and clinical relevance.

\section*{Acknowledgments}
We acknowledge Analog Devices for research support. Dr. Emad Boctor was supported by National Science Foundation CAREER Award [1653322]. Yixuan Wu was supported by National Institutes of Health Graduate Partnerships Program (GPP) Fellowship.

\bibliographystyle{IEEEtran}
\bibliography{RA-L_manuscript}

\vfill

\end{document}